\documentclass{article}

\usepackage[nonatbib,final]{neurips_2024}

\usepackage[utf8]{inputenc} %
\usepackage[T1]{fontenc}    %
\usepackage{hyperref}       %
\usepackage{url}            %
\usepackage{booktabs}       %
\usepackage{amsfonts}       %
\usepackage{nicefrac}       %
\usepackage{microtype}      %
\usepackage{xcolor}         %

\usepackage[noend]{algpseudocode}

\usepackage{amssymb}
\usepackage{optidef}

\usepackage[english]{babel}
\usepackage{blindtext}
\usepackage{xspace}
\usepackage{amsmath}
\usepackage[normalem]{ulem}
\usepackage{caption}
\usepackage{graphicx}
\usepackage{booktabs}
\usepackage{arydshln}
\usepackage{xurl}
\usepackage{subcaption}
\usepackage{wrapfig}
\usepackage[toc,title]{appendix}
\usepackage{afterpage}
\usepackage{tikz}
\usepackage{enumitem}
\usepackage{stfloats}
\usepackage[ruled,linesnumbered]{algorithm2e}

\SetCommentSty{mycommfont}
\usepackage{comment}
\usepackage{algorithmicx}
\usepackage{algpseudocode}
\usepackage{multirow}

\usepackage{xspace}
\usepackage{mfirstuc} %
\usepackage{longtable}
\usepackage{listings}
\usepackage{xcolor}

\usepackage[most]{tcolorbox} %
\usepackage{graphicx} %
\usepackage{multicol} 

\usepackage[affil-it]{authblk} 
\usepackage{hyperref}
\usepackage{tikz}
\usepackage{graphicx}
\usepackage{subcaption}

\newcommand{\sys}{\textit{Eagle}\xspace}
\newcommand{\sysg}{\textit{Eagle-Global}\xspace}
\newcommand{\sysl}{\textit{Eagle-Local}\xspace}

  {\begin{list}{\labelitemi}{\itemsep2pt \topsep2pt \parsep0.00in
  \partopsep=0pt \leftmargin1.2em}}%
  {\end{list}}

\let\latexusecounter=\usecounter
\def\compactsortof{\itemsep=2pt \topsep=2pt \parsep=0.00in \partopsep=0pt
\leftmargin=1.2em}

\newcommand{\BULLET}{\vspace{+.00in} \noindent $\bullet$ \hspace{+.00in}}

\newcommand{\mysection}[1]{\vspace{-.09in}\section{#1}\vspace{-.03in}}
\newcommand{\mysubsection}[1]{\vspace{-.09in}\subsection{#1}\vspace{-.06in}}

\title{Eagle: Efficient Training-Free Router for Multi-LLM Inference}

\author{%
  Zesen Zhao\thanks{Equal contribution.} , Shuowei Jin\protect\footnotemark[1] , Z. Morley Mao \\
  University of Michigan\\
  \texttt{\{hymanzzs, jinsw, zmao@umich.edu\}} \\
}

\begin{document}
\maketitle

\begin{abstract}
    The proliferation of Large Language Models (LLMs) with varying capabilities and costs has created a need for efficient model selection in AI systems. LLM routers address this need by dynamically choosing the most suitable model for a given query based on task requirements and budget constraints. However, existing routers face challenges in scalability and real-time adaptation, particularly in high-volume online environments. We present \sys, a novel LLM routing approach that combines global and local ELO ranking modules to overcome these limitations. By evaluating both general and specialized LLM abilities, \sys provides a scalable, training-free solution that enhances model selection quality while reducing computational overhead. Our experiments across multiple datasets show \sys consistently outperforms baseline methods, with improvements of up to 23.52\% in Area Under Curve (AUC) scores. 
    Moreover, \sys demonstrates remarkable efficiency, requiring only 1/20 of baseline methods' time for initialization and 100-200x faster incremental updates in online scenarios, making it well-suited for dynamic, high-volume online serving environments.
\end{abstract}

\mysection{Introduction}

Large language models (LLMs) have demonstrated extraordinary performance across numerous tasks, from daily tasks like commonsense reasoning and question answering, to advanced tasks like code generation and mathematical problem-solving ~\cite{brown2020language,chowdhery2023palm,wei2022chain,jin2024adaptive}.

Two primary strategies have emerged to improve LLM performance across various tasks. The first is scaling up~\cite{kaplan2020scaling}: increasing model size and training data to boost overall performance across a wide range of tasks. However, this method comes with the tradeoff of significantly higher inference costs due to the larger number of parameters. The second approach is specialization: developing domain-specific LLMs that can achieve performance comparable to larger models within a specific domain while maintaining a much smaller model size. For instance, CodeQwen~\cite{qwen} demonstrates GPT-4-level code generation capabilities with only a 7B parameter model~\cite{evalplus}.

These developments have resulted in a diverse ecosystem of LLMs, each with its own specialties and associated inference costs. To leverage the varying qualities and costs of these models effectively, researchers have proposed the concept of a "router" – a system designed to select the optimal model for a given query based on content and cost constraints~\cite{ong2024routellm, hu2024routerbench}. Compared to traditional single-model query procedures, a router enables users to obtain the highest quality answer within their budget. The primary goal of the router is to predict quality rankings, as the cost of querying each model is fixed. LLM service providers are beginning to integrate routers into their systems, with OpenAI recently releasing an "Auto" feature on the ChatGPT website to dynamically choose between their models (GPT-4o, GPT-4o mini, o1-preview) based on task requirements and performance characteristics.

However, designing a practical LLM router for online serving systems presents several challenges:

\BULLET \textbf{Scalability and Real-time Adaptation:} With millions of requests per second, the router must efficiently process and route queries while continuously adapting to new information. Traditional machine learning approaches often struggle with the computational overhead of frequent retraining.

\BULLET \textbf{Incomplete Feedback Data:} User feedback in online systems is often limited to pairwise comparisons between two responses. Extracting meaningful global rankings from this partial information is a non-trivial task that many existing methods fail to address adequately.

\BULLET \textbf{Balancing Accuracy and Efficiency:} While high prediction accuracy is crucial for optimal routing, achieving this accuracy must not come at the cost of system responsiveness. Striking the right balance between these often-competing goals remains a significant challenge in LLM routing.

Existing approaches to LLM routing have limitations. Routerbench~\cite{hu2024routerbench} proposes using traditional machine-learning methods like MLP and KNN to predict quality rankings, but these require heavy retraining and cannot utilize real-time user feedback. 
RouteLLM~\cite{ong2024routellm} proposes a similarity-weighted ranking approach for training-free prediction. While effective for binary routing decisions, their method is limited to scenarios involving only two models and does not address cases with $n \geq 3$ models (where $n$ denotes the number of models).

To address these challenges, we present~\sys, a novel approach to LLM routing with the following key contributions:

\BULLET \sys achieves the most accurate prediction of model quality rankings by designing a global and local ELO module, ensuring the highest-quality answers within a given budget. Our experiments show that \sys outperforms baseline methods across all datasets, with improvements of 23.52\% over SVM, 5.14\% over KNN, and 4.73\% over MLP in terms of Area Under Curve (AUC) scores.

\BULLET \sys is a training-free router that is significantly more efficient than traditional machine learning-based routers when adapting to newly collected data. Our experiments demonstrate that \sys requires only 4.8\% of the training time of baseline methods for initial setup, and a mere 0.5-1\% of the updating time for incremental data updates.

\mysection{\sys Architecture and Design}

\begin{figure}[t!]
    \centering
    \includegraphics[width=0.9 \linewidth]{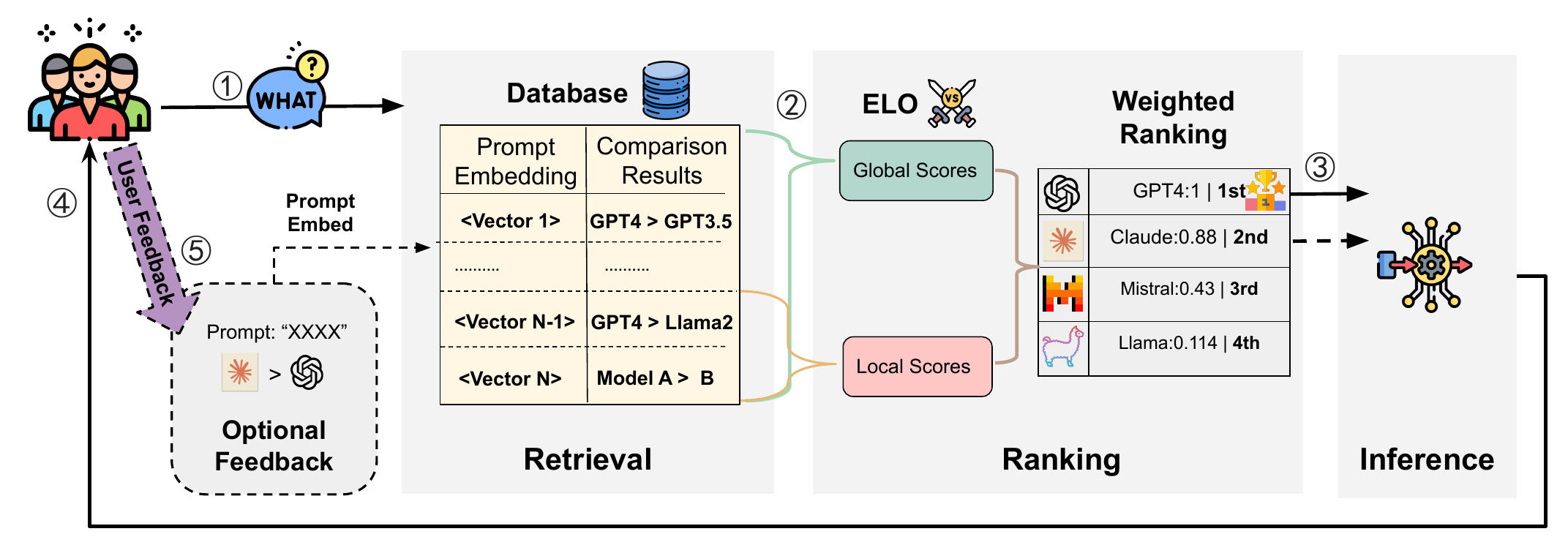}
    \caption{\sys workflow: \textcircled{1} User request submission. \textcircled{2} Retrieval of relevant historical data. \textcircled{3} LLM quality ranking and selection within budget. \textcircled{4} Response generation and delivery. \textcircled{5} Optional secondary model comparison and feedback collection.}
    \vspace{-0.6cm}
    \label{fig:eagle-diagram}
\end{figure}

We present the workflow of \sys in Figure~\ref{fig:eagle-diagram}. \sys is designed to optimize the selection of Large Language Models (LLMs) based on user requirements and historical performance. The system maintains a comprehensive vector database containing embeddings of previous input prompts along with their corresponding user feedback. When a user submits a request, \sys leverages this database to retrieve relevant historical information and generates a predicted response quality ranking for various available LLM models. Considering the user's budget constraints, \sys then selects the highest quality model within the specified budget. The user's query is then routed to the selected LLM for inference. After the target LLM generates the response, the result is returned to the user. To gather additional feedback, \sys may optionally select another model from the list and ask the user to compare the quality of the two responses.

\mysubsection{\sys Design}

The key idea behind \sys's design for accurately predicting different models' response 
quality ranking is based on two main insights:

1. \textbf{Holistic Ability Assessment}: Each LLM possesses two types of abilities:
   a) General Ability: A model's overall performance across diverse tasks, captured by analyzing its behavior on the entire dataset.
   b) Specialized Ability: Task-specific proficiency, identified through performance patterns on similar historical queries.

2. \textbf{Efficient Feedback Integration} Since user feedback typically only provides a preference between two models, obtaining a complete ranking of all models from users is challenging. To address this, we employ the ELO\cite{elo1967uscf} ranking method, which transforms sparse pairwise comparisons into a comprehensive ranking system, maximizing the value of each user interaction.

To implement these principles, \sys incorporates two core modules:

- \sysg: Evaluates models' general ability using the entire historical dataset.

- \sysl: Assesses specialized abilities by analyzing performance on similar past queries.

Both modules leverage the ELO algorithm to construct full model rankings from partial feedback data. By combining both \sysg and \sysl, \sys achieves a more accurate and nuanced prediction of model performance, adapting to both general trends and query-specific requirements.

\mysubsection{Details of \sys}

The ELO algorithm is a rating system used to calculate the relative skill levels of players (or models) by updating their scores based on the outcomes of pairwise comparisons. The ELO rating for a model is updated after each match using the following formula:

\begin{equation}
R' = R + K \times (S - E)
\end{equation}

Where \( R \) is the current rating, \( R' \) is the updated rating, \( S \) is the actual score (1 for win, 0.5 for draw, 0 for loss), \( K \) is a constant determining the sensitivity of rating changes, higher  \( K \) results in larger adjustments, and \( E \) represents the expected score or the predicted probability of a player winning a match based on their 
current ratings and is calculated using the rating difference between two players:
\vspace{-0.1in}
\begin{equation}
E = \frac{1}{1 + 10^{\frac{R_{\text{opponent}} - R}{400}}}
\end{equation}

For \sysg, we calculate the average ELO rating across all pairwise feedback information in the database. When new feedback data is collected, we can efficiently update \sysg by performing ELO calculations on the new data only, eliminating the need for retraining.

To operate \sysl, we first utilize a vector database to retrieve the \( N \) nearest neighbors of user feedback based on cosine semantic similarity using the prompt embedding vector. We initialize the local ELO scores for each model using the global ELO scores as background knowledge. Then, we use the retrieved local feedback to update the local ELO scores for each query.

With these two scores, we compute the weighted sum of the global and local ELO scores using the following equation: $\text{Score}(X) = P \times \text{Global}(X) + (1 - P) \times \text{Local}(X)$. This combined score provides a comprehensive quality ranking of the models, accounting for both their general and specialized abilities. \sys then selects the highest-ranked model that falls within the user's specified budget constraints, ensuring an optimal balance between performance and cost-efficiency.

\mysection{Evaluation}
In this section, we present a series of experiments designed to evaluate the effectiveness and efficiency of \sys. All of our experiments were conducted using the RouterBench dataset~\cite{hu2024routerbench}. Detailed information about the experimental setup, including hardware specifications, model parameters, and baseline configurations, can be found in Appendix~\ref{appendix:exp_setup}. 
We also perform an ablation study to understand the effects of \sysg and \sysl on the final performance, which is detailed in Appendix~\ref{appendix:ablation}.

\subsection{Overall Performance}

\begin{figure}[t!]
    \centering
    \begin{subfigure}[b]{0.5\textwidth}
        \includegraphics[width=\textwidth]{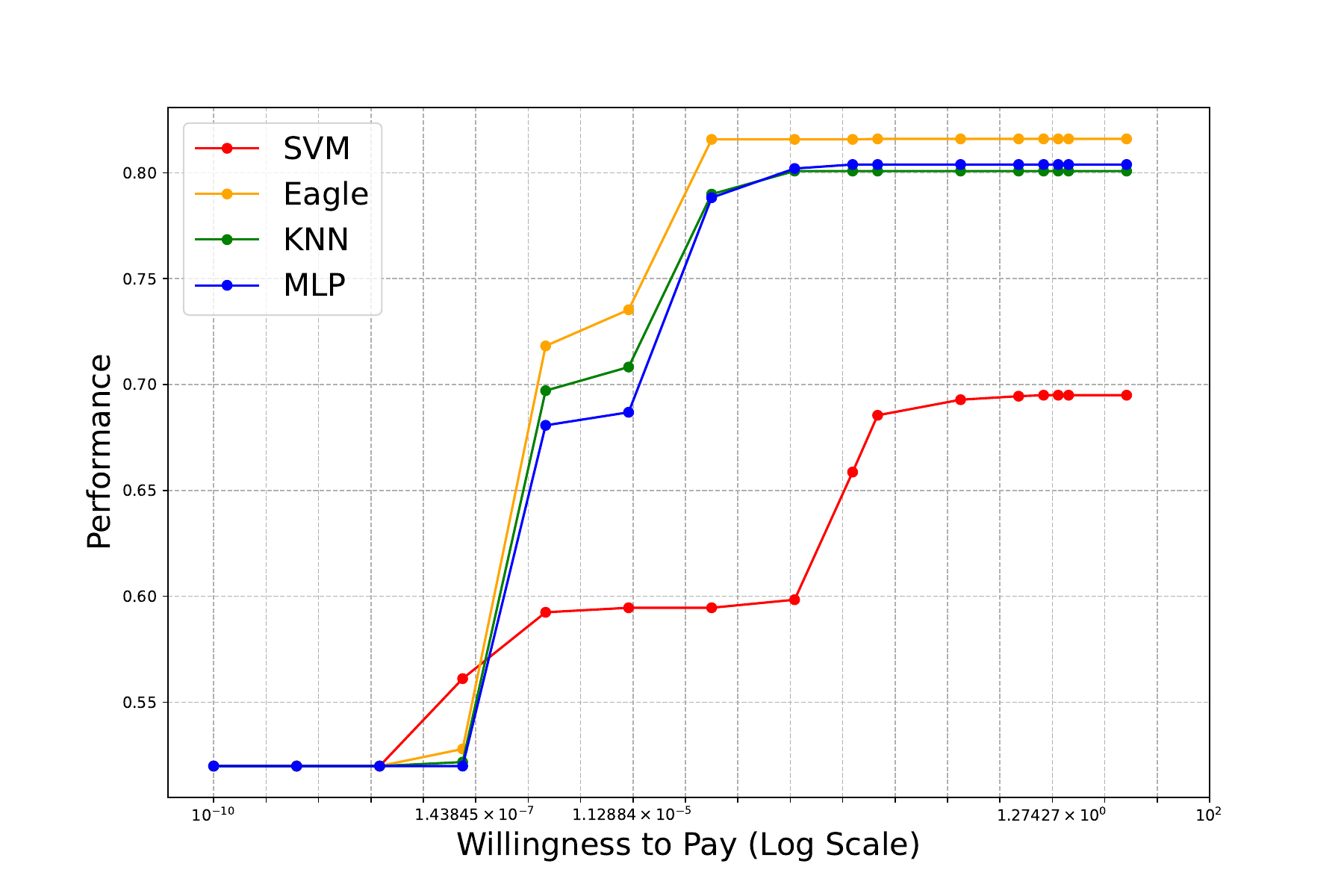}
        \caption{Router's performance with budget on MMLU dataset.}
        \label{fig:3a}
    \end{subfigure}
    \hfill
    \begin{subfigure}[b]{0.49\textwidth}
        \includegraphics[width=\textwidth]{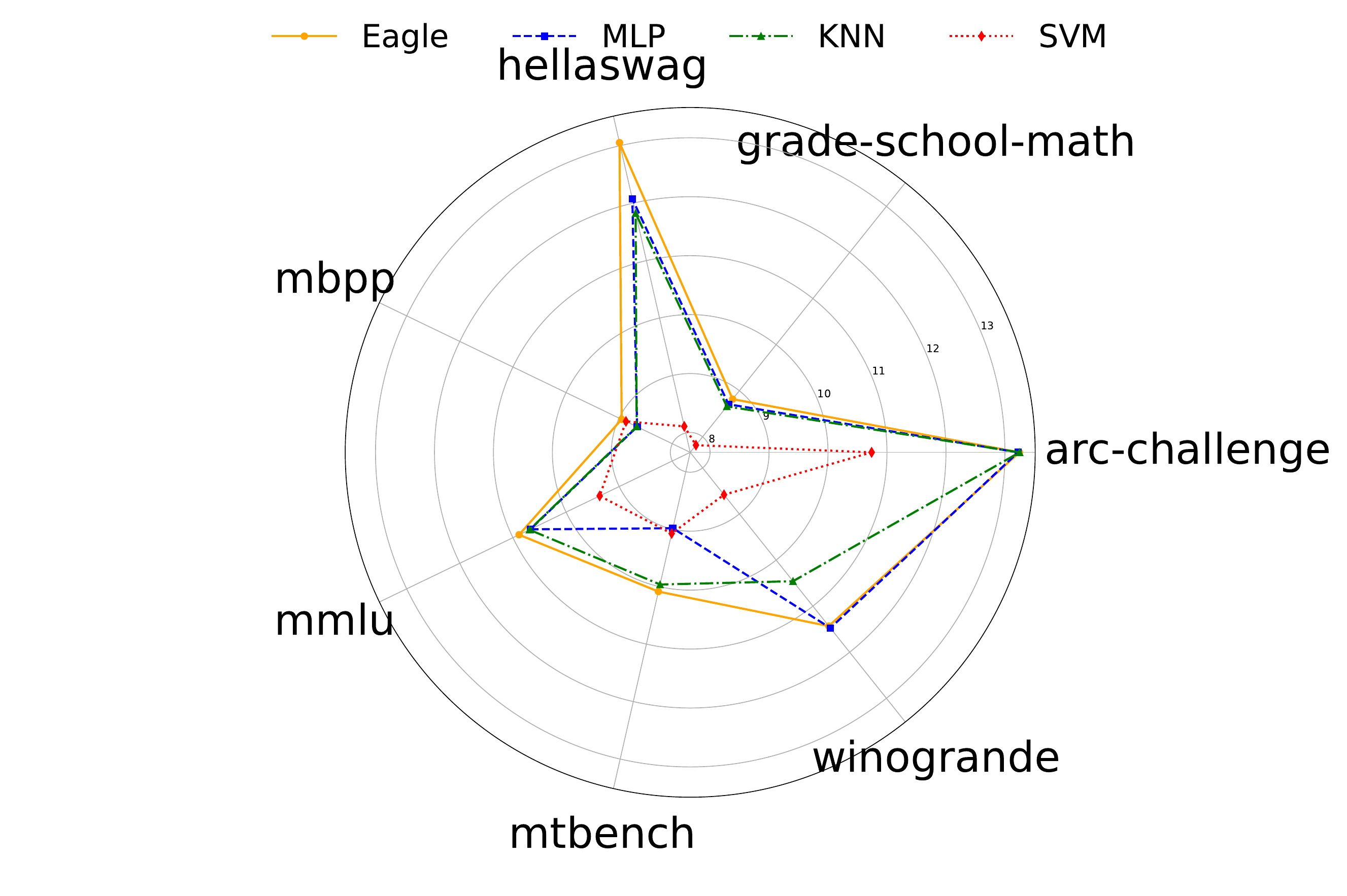}
        \caption{Area Under Curve across seven datasets.}
        \label{fig:3b}
    \end{subfigure}
    \vspace{-0.2in}
    \caption{Comparison of Baseline Models with Eagle.\vspace{-0.1in}}
    \label{fig:overall_performance}
\end{figure}

We evaluated the performance of our method against baseline methods across all datasets in RouterBench, including MMLU\cite{mmlu}, Hellaswag\cite{hell}, GSM8K\cite{gsm}, ARC Challenge\cite{arc}, Winogrande\cite{wing}, MBPP\cite{mbpp}, and MT-Bench\cite{mt}. Figure~\ref{fig:3a} illustrates the willingness-to-pay versus performance graph for each router on the MMLU dataset. As observed, \sys consistently outperforms other methods across various willingness-to-pay levels.

To comprehensively compare performance across all datasets, we calculated the area under the curve (AUC) using the trapezoidal rule. The AUC serves as a metric to evaluate a router's average performance across all cost scenarios. Figure~\ref{fig:3b} presents a radar graph comparing the AUC of \sys and baseline methods across the seven datasets. \sys demonstrates superior performance across all datasets. Summing up the AUC scores, we find that \sys achieves a 23.52\% improvement over SVM, a 5.14\% improvement over KNN, and a 4.73\% improvement over MLP.

\begin{figure}[t]
    \centering
    \begin{subfigure}[b]{0.5\textwidth}
        \centering
        \begin{tabular}{|c|c|c|c|}
            \hline
            Models & 70\% & 85\% & 100\% \\ \hline
            KNN & 176.3 & 180.6 & 193.4 \\ \hline
            MLP & 248.3 & 253.3 & 260.2 \\ \hline
            SVM & 114.7 & 143.0 & 150.5 \\ \hline
            \sys & \textbf{8.0} & \textbf{1.4} & \textbf{1.5} \\ \hline
        \end{tabular}
        \caption{Training time (seconds) for models at different data stages. 70\%: initial training; 85\% and 100\%: incremental adding data simulating user feedback collection.}
        \label{tab:retraining_time_table}
    \end{subfigure}
    \hfill
    \begin{subfigure}[b]{0.45\textwidth}
        \centering
        \includegraphics[width=0.9\textwidth]{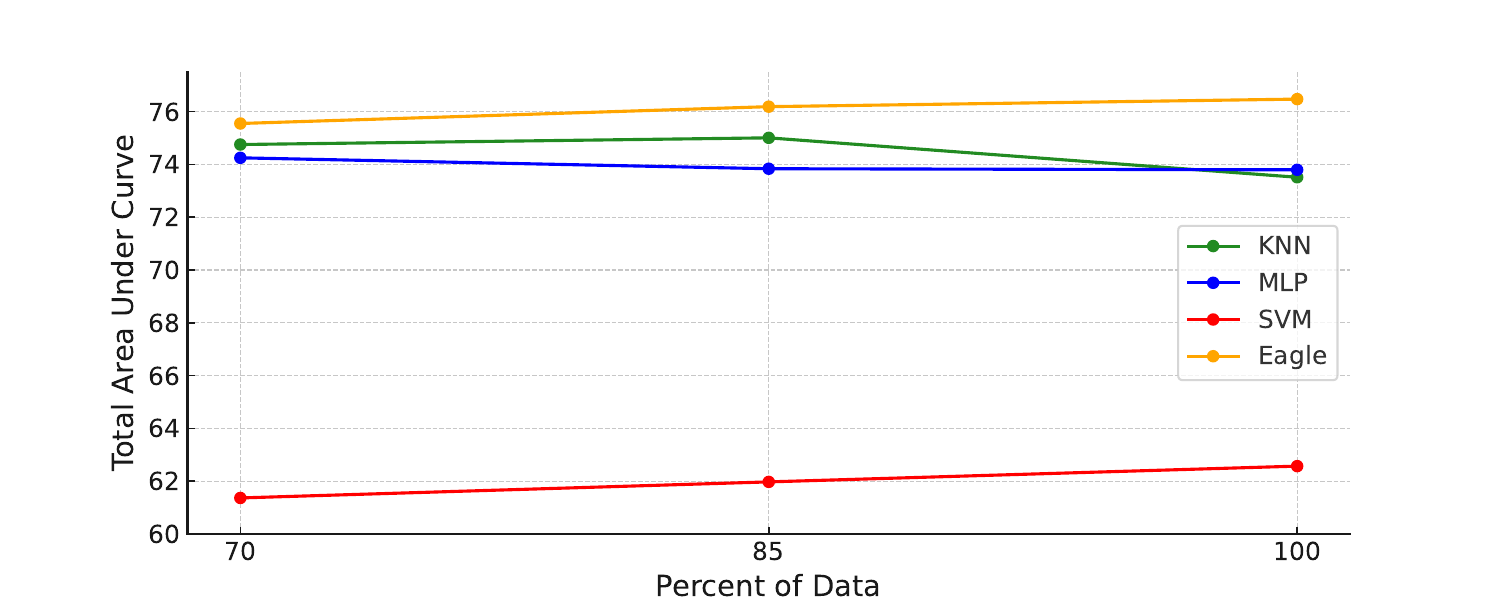}
        \caption{Routers' performance on test set when incrementally using more data.}
        \label{fig:incremental_update}
    \end{subfigure}
    \caption{Comparison of training time and quality \vspace{-0.2in}}
    \label{fig:combined}
\end{figure}

\subsection{Online Adaptation Efficiency and Quality}
In this section, we evaluate the efficiency and effectiveness of our method compared to baseline models in an online serving scenario. We simulate this by initially training all methods on 70\% of the training data, then assessing the retraining time and performance when each additional 15\% of data is introduced. Table~\ref{tab:retraining_time_table} illustrates the efficiency (training time) of \sys compared to the baseline methods. Even at the initial full data training stage, \sys's global ELO score initialization is significantly more efficient, using only 4.8\% of the training time required by baseline methods. This efficiency stems from \sys's approach of updating global scores once, rather than iteratively optimizing the model. The efficiency gap widens as new data is introduced, with \sys requiring only 0.5-1\% of the baseline updating time for each 15\% increment of new data.

Importantly, this efficiency does not come at the cost of performance. We evaluated \sys against baseline methods across seven datasets and calculated the summed Area Under the Curve (AUC) metric. As demonstrated in Figure~\ref{fig:incremental_update}, our results show that \sys consistently outperforms other methods across all dataset settings, demonstrating both superior efficiency and effectiveness. At 70\% of the data, \sys achieves an average quality improvement of 8.65\% across the three baseline routers. This improvement increases to 9.21\% at 85\% data and 9.92\% at 100\% data, demonstrating consistently superior performance across varying data volumes.

\bibliography{reference}
\bibliographystyle{plain}

\clearpage

\appendix
\mysection{Experimental Setup}
\label{appendix:exp_setup}

The experiments were performed on a server equipped with a Ryzen 3700X CPU, an RTX 4070 GPU, and 32GB of RAM. We utilized the \texttt{stella\_en\_1.5B\_v5}~\cite{stella} model for text embedding. The dataset was split into 70\% for training and validation, with the remaining 30\% used for testing.

\subsection{Model Parameters}
For \sys, we set the following parameters:
\begin{itemize}
    \item \( P = 0.5 \)
    \item \( N = 20 \)
    \item \( K = 32 \)
\end{itemize}

\subsection{Baseline Configurations}
For the baseline methods, we used the following configurations:
\begin{itemize}
    \item Common settings: neighbor size of 40 and cosine similarity as the distance function
    \item MLP: two layers with a hidden size of 100 and ReLU activation
    \item KNN: 40 nearest neighbors with cosine similarity
    \item SVM: LinearSVR with epsilon set to 0.0
\end{itemize}

\mysection{Ablation Studies for \sys}
\label{appendix:ablation}

\begin{figure}[htbp]
    \centering
    \begin{subfigure}[b]{0.45\textwidth}
        \includegraphics[width=\textwidth]{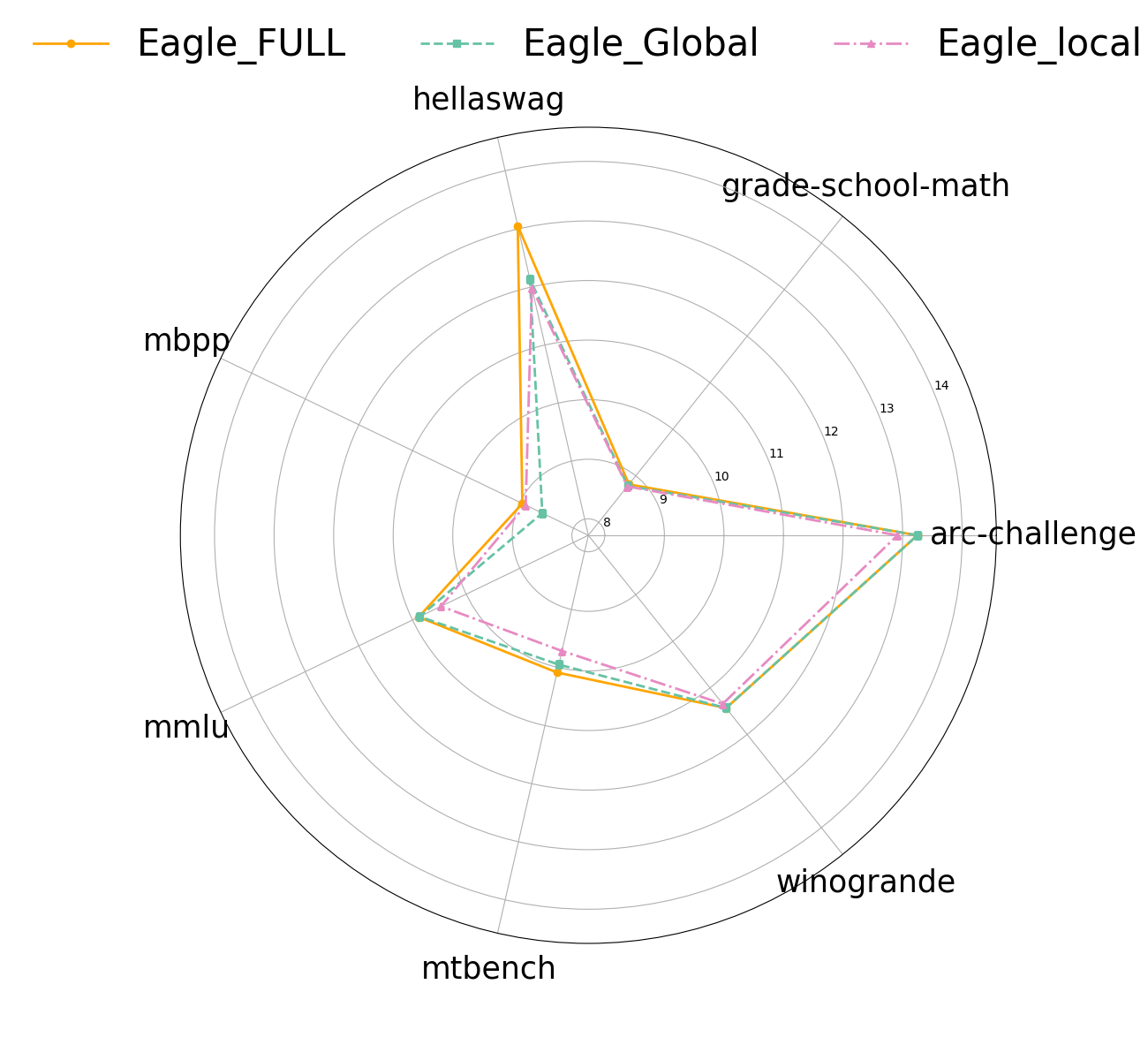}
        \caption{Performance comparison of \sysg-only, \sysl-only, and \sys.}
        \label{fig:3}
    \end{subfigure}
    \hfill
    \begin{subfigure}[b]{0.45\textwidth}
        \includegraphics[width=\textwidth]{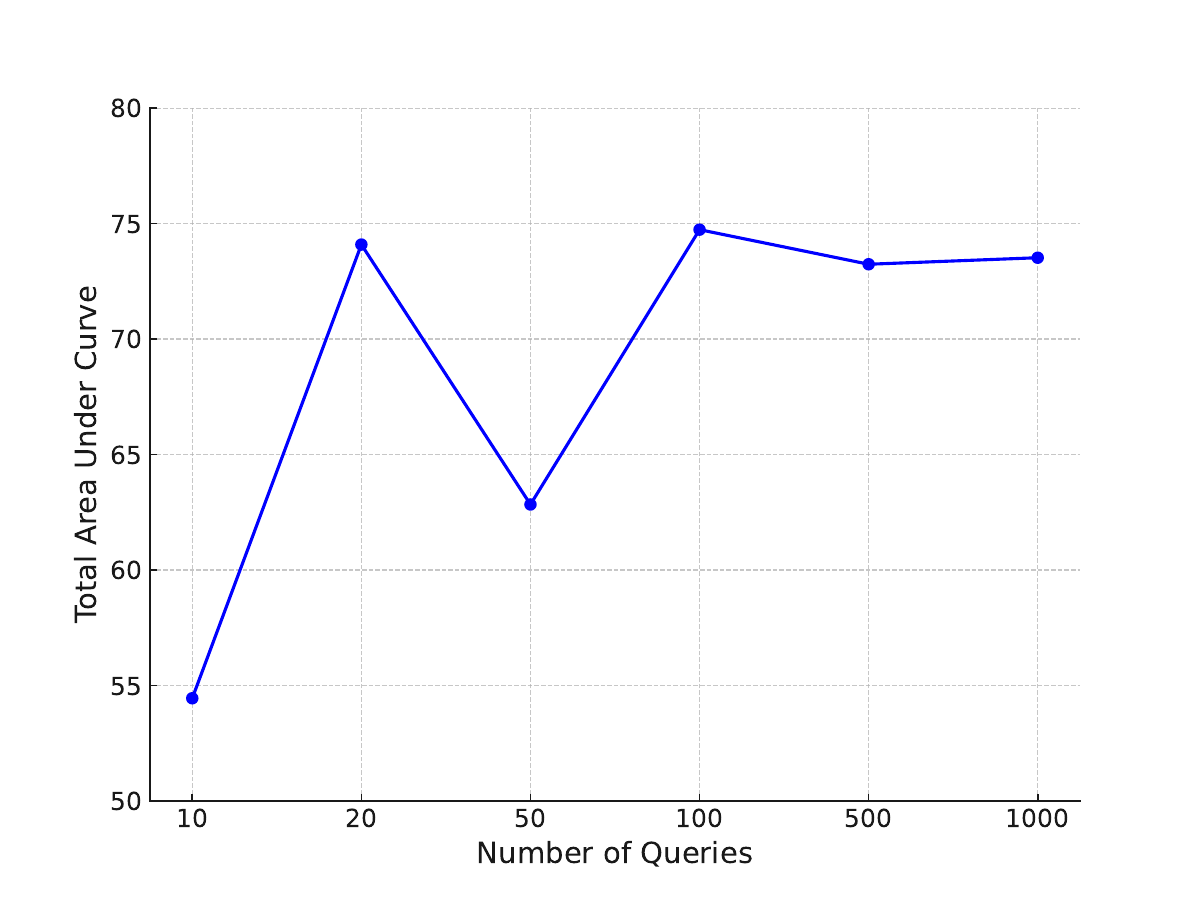}
        \caption{Effect of local neighbor size ($N$) on \sysl performance.}
        \label{fig:4}
    \end{subfigure}
    \caption{Ablation studies for \sys components and parameter sensitivity.}
    \label{fig:combined_figures}
\end{figure}

To further understand the effectiveness of \sys, we conducted ablation studies comparing the performance of \sys with its individual components: \sysg and \sysl. Figure~\ref{fig:3} illustrates the results of these comparisons.

Our findings reveal that neither \sysg nor \sysl alone can achieve optimal performance. \sysg, while effective in capturing global information, lacks the ability to focus on different models' specialized capabilities. Conversely, \sysl excels at identifying model-specific strengths but may be biased due to limited sample sizes. The combination of \sysg and \sysl in \sys leverages the strengths of both approaches, resulting in superior overall performance.

We also investigated the impact of the local neighbor size ($N$) on \sysl performance. As shown in Figure~\ref{fig:4}, we observed that when $N=10$, \sysl lacks sufficient information to make accurate predictions. However, increasing $N$ beyond a certain point yields diminishing returns. Our experiments indicate that $N=20$ provides the optimal balance between performance and computational efficiency.

\mysection{Related Works}

Large language models (LLMs) have demonstrated exceptional capabilities across a wide range of tasks. However, training and serving a single massive model is both costly and inefficient. Additionally, recent findings show that larger models do not consistently outperform smaller or specialized LLMs for all tasks. To address these issues, researchers are exploring multi-LLM approaches to enhance system performance while maintaining cost efficiency.

\textbf{Mixture-of-Experts (MoE) and Ensemble Learning} are two pivotal techniques for optimizing multi-LLM systems by leveraging multiple models to improve both performance and efficiency. Ensemble Learning, seen in systems like LLM Blender\cite{jiang2023llm} and Blending Is All You Need\cite{lu2024blendingneedcheaperbetter}, combines outputs from multiple models to enhance accuracy and robustness, albeit often at the cost of increased computational overhead. In contrast, MoE\cite{shazeer2017outrageouslylargeneuralnetworks} activates only a subset of experts for each task, reducing computational demands by using only the most relevant models. While both approaches aim to boost LLM performance through the use of multiple models, MoE emphasizes scalability and resource efficiency, whereas Ensemble Learning focuses on robustness by combining model outputs. Nonetheless, challenges such as increased complexity in ensemble methods and potential inefficiencies in expert selection for MoE remain.

\textbf{Router-based methods}, including Route LLM\cite{ong2024routellm}, PolyRouter\cite{stripelis2024polyrouter}, hybrid LLM\cite{ding2024hybridllmcostefficientqualityaware}, and Intelligent Router for LLM Workloads\cite{jain2024intelligent}, strive to enhance efficiency by dynamically routing queries to the most suitable model. These methods intelligently allocate tasks based on factors like task complexity, model performance, and system load, minimizing unnecessary computation and optimizing resource utilization. Route LLM focuses on matching queries to the most capable model, PolyRouter balances performance with cost, hybrid LLM tries to predict query complexity and route to most suitable models rather than singleton superior LLM, and Intelligent Router applies workload-aware scheduling to maximize throughput under heavy loads. While these approaches improve efficiency, they often introduce complexity in designing effective routing algorithms and managing real-time coordination among multiple models. To facilitate fair comparisons between routing strategies, benchmarks like RouterBench\cite{hu2024routerbench} and Large Language Model Routing with Benchmark Datasets\cite{shnitzer2023largelanguagemodelrouting} provide standardized metrics that assess performance, efficiency, and resource consumption.

However, many current systems still increase computational costs or require additional training when processing new user data. No existing approach can adaptively update route designation in real time based on a user’s recent queries.

\end{document}